\documentclass[11pt]{article}

\usepackage[preprint]{acl}

\usepackage{times}
\usepackage{latexsym}
\usepackage{amsmath}
\usepackage{booktabs}
\usepackage{arydshln}
\usepackage{multirow}
\usepackage{placeins}
\usepackage{CJKutf8}


\usepackage[T1]{fontenc}

\usepackage[utf8]{inputenc}

\usepackage{microtype}

\usepackage{inconsolata}

\usepackage{graphicx}

\title{RP-OPSD: Reasoning-Pivot-Guided On-Policy Self-Distillation for Multilingual Reasoning Transfer}

\author{Xinye Wang, Junxiao Liu, {Shujian Huang}$^{\dagger}$\\
  National Key Laboratory for Novel Software Technology, Nanjing University \\
  \texttt{\{xinye.wang, junxiao.liu\}@smail.nju.edu.cn, huangsj@nju.edu.cn} }

\begin{document}
\maketitle
\begin{abstract}
Multilingual reasoning transfer is crucial for extending reasoning capabilities of large language models (LLMs) beyond high-resource languages. 
On-policy self-distillation (OPSD) and its variants have emerged as a promising paradigm, providing dense token-level supervision on student-generated rollouts, yet their objectives do not explicitly prioritize reasoning signals most critical to cross-lingual transfer. 
We characterize that target-language reasoning comprises the generation of both surface text and \textit{reasoning pivots}, which are decisions that advance or redirect the reasoning process and shape subsequent inference. This motivates concentrating privileged distillation around such pivots. 
We therefore propose RP-OPSD, Reasoning-Pivot-guided On-Policy Self-Distillation, using the distributional shift between matched teacher views with and without an English reference solution as an operational proxy to guide privileged distillation and reference anchoring. 
Experiments on mathematical reasoning benchmarks covering 17 languages and multiple difficulty levels show that our method outperforms strong multilingual reasoning baselines and OPSD variants. 
Further analysis reveals that RP-OPSD concentrates privileged distillation on reasoning-control and problem-condistioned state-update tokens, while downweighting it for tokens that mainly support surface realization.
Our code is available at \url{https://github.com/NJUNLP/RP-OPSD}.
\end{abstract}

\section{Introduction}

\begin{figure}[t]
  \centering
  \includegraphics[width=\linewidth]{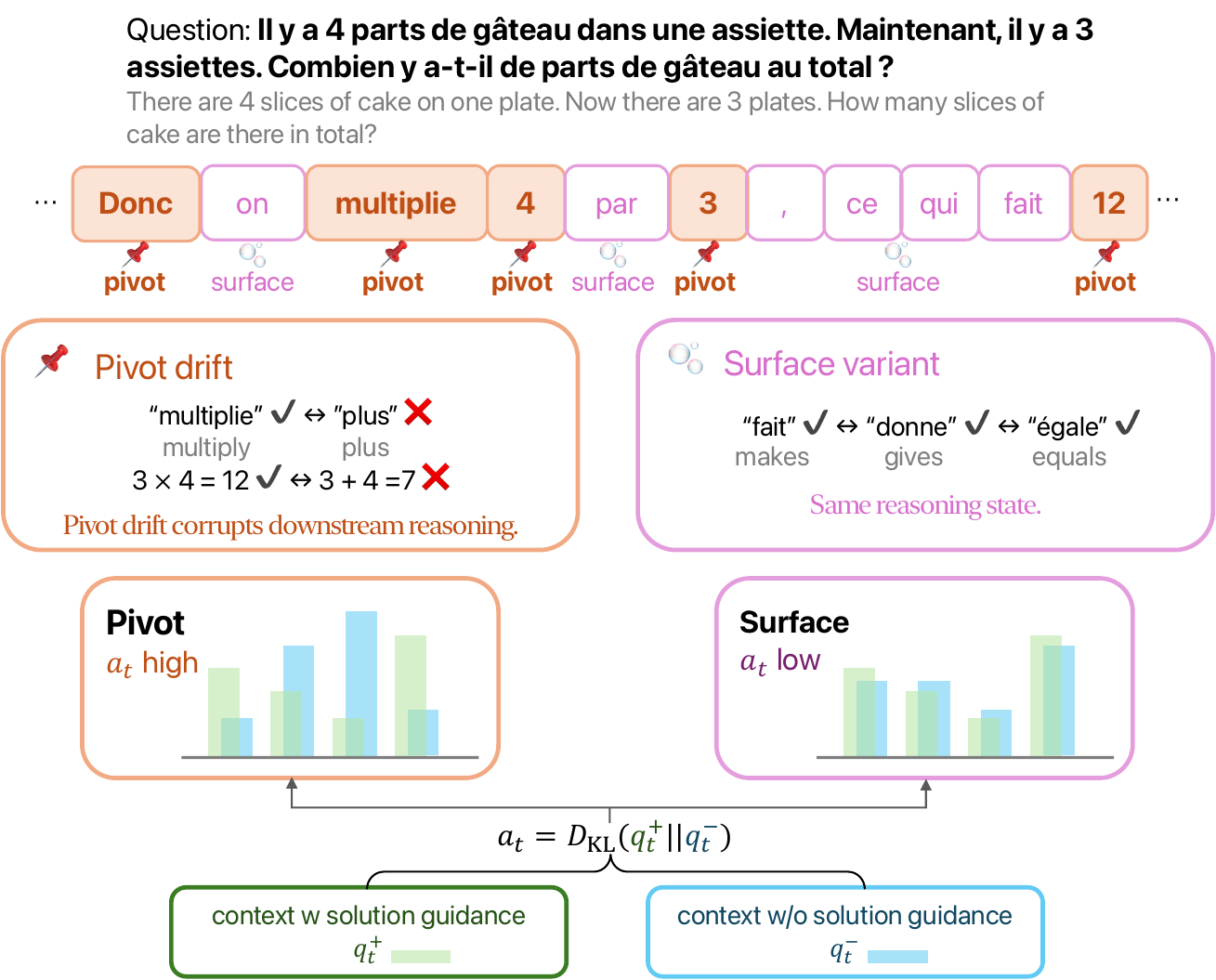}
  \caption{Motivation of reasoning-pivot-guided transfer. Surface variants can preserve the same reasoning state, while pivot drift corrupts downstream reasoning. RP-OPSD uses the distribution gap between the solution-conditioned view $q_t^{+}$ and the ablated view $q_t^{-}$ to identify reasoning-sensitive tokens. The Frence CoT means "So we multiply 4 by 3, which makes 12".}
  \label{fig:motivation}
\end{figure}

Large language models (LLMs) exhibit substantial multi-step reasoning ability \citep{wei2023chainofthoughtpromptingelicitsreasoning, Guo_2025}.
In multilingual settings, however, the accessibility of this ability remains highly uneven among different languages \citep{shi2022languagemodelsmultilingualchainofthought, ahuja-etal-2023-mega}.
Due to the severe imbalance in multilingual training data \citep{nguyen-etal-2024-democratizing, nag-etal-2025-efficient}, models often fail to reproduce in non-dominant or low-resource languages the reasoning behaviors they exhibit in English.
This disparity makes multilingual reasoning transfer a central problem: enabling high-resource reasoning capabilities to be reliably elicited, transferred, and realized in low-resource linguistic contexts.

Existing adaptation strategies address this gap only partially.
Supervised fine-tuning (SFT) on translated rationales supplies dense training signals  \citep{lai-nissim-2024-mcot, barua2026longchainofthoughtreasoninglanguages}, but translated rationales are off-policy and can contain translation artifacts that deviate from the model's own low-resource reasoning trajectories.
Reinforcement learning (RL) methods such as GRPO \citep{shao2024deepseekmathpushinglimitsmathematical} avoid fixed rationales, yet correctness-only rewards are sparse and sequence-level, assigning credit only after a final answer is verified \citep{zhang-etal-2026-think, liu2026trit}.
Such rewards provide limited pressure on the language fidelity and intermediate reasoning structure of target-language chains of thought (CoT)  \citep{huang2026tapotranslationaugmentedpolicy, wang-etal-2025-demystifying-multilingual, tran2025exploitingtreestructurecredit}.

Recently, on-policy self-distillation (OPSD) has emerged as a viable alternative: it provides dense token-level supervision directly on student-generated trajectories \citep{zhao2026selfdistilledreasoneronpolicyselfdistillation}. COPSD adapts this paradigm to multilingual reasoning using a privileged cross-lingual teacher context, but its objective assigns the same weight to every generated position \citep{liu2026copsd}. This uniformity may dilute the transferable reasoning signal across positions devoted primarily to target-language realization. We argue that a target-language CoT interleaves language-specific surface realization with \emph{reasoning pivots}: local decisions that determine how the latent solution state evolves, such as selecting a subgoal, applying an operation, binding a variable, or making an inference transition. As Figure~\ref{fig:motivation} illustrates, surface variants can preserve the same reasoning state, whereas pivot drift can corrupt downstream reasoning even when the continuation remains fluent.

This distinction raises a central question: which response tokens should receive stronger privileged distillation. That is, how to locate reasoning pivots? Existing OPSD variants use different proxies: EGRSD and CL-EGRSD weight positions according to teacher-confidence signals \citep{ke2026respectingselfuncertaintyonpolicyselfdistillation}, whereas TRACE routes distillation to annotator-identified key/error spans \citep{wang2026tracedistillingmatterstokenrouted}. However, neither criterion directly identifies transferable pivots in multilingual CoTs: teacher confidence need not reflect cross-lingual reasoning relevance, while critical-span annotations are difficult to obtain consistently across languages.

An intuitive annotation-free alternative is to prioritize tokens by raw teacher-student KL, yet recent work suggests that such disagreement is a coarse selection signal: a large distillation residual need not indicate a reasoning-critical position \citep{xu2026tiptokenimportanceonpolicy,wang2026disagreementlearnabletokenteachability,huang2026skillconditionedgatedselfdistillationllm}. For locating reasoning pivots, a more task-aligned signal should instead quantify the incremental effect of privileged reasoning evidence under a controlled information state.

To this end, we propose RP-OPSD: Reasoning-Pivot-Guided On-Policy Self-Distillation for multilingual reasoning transfer. 
On student-generated target-language rollouts, RP-OPSD turns the distributional shift between teacher views into a token-level Reasoning-Pivot Transfer (RPT) gate. The gate routes full-distribution privileged distillation toward high-sensitivity positions and frozen-reference anchoring elsewhere, concentrating transfer where solution access most affects the reasoning continuation while preserving surface realization.

Across 17 languages and two mathematical-reasoning benchmarks, RP-OPSD outperforms strong multilingual reasoning baselines and OPSD variants.
Further analyses reveal reasoning-control transitions and problem-conditioned state updates among high-gate positions, in contrast to surface target-language and symbolic realization at low-gate positions.

\section{Methodology}

Along an on-policy target-language rollout (Section~\ref{sec:method_notations}), RP-OPSD localizes candidate reasoning pivots by contrasting two teacher views that share the bilingual question and rollout prefix but differ only in access to the English reference trace (Section~\ref{sec:method_teacher_views}).
The resulting gate (Section~\ref{sec:method_rpt_gate}) continuously balances privileged distillation with target-language reference anchoring (Section~\ref{sec:method_objective}).
\subsection{Notations}
\label{sec:method_notations}

For each training instance, let $x^\ell$ denote the low-resource question, $x^h$ its English translation, and $s^h$ an English reference reasoning trace.
The trainable policy is denoted by $\pi_\theta$.
The student rollout is generated on policy from the low-resource input:
\[
y_{1:T}\sim \pi_\theta(\cdot \mid x^\ell).
\]

At response position $t$, the student next-token distribution is
\[
p_t
=
\pi_\theta(\cdot \mid x^\ell, y_{<t}).
\]
Only this distribution receives gradients.

RP-OPSD does not introduce a separate teacher network.
Following OPSD, teacher distributions are stop-gradient evaluations of the same policy under privileged information.
We use $\operatorname{sg}[\cdot]$ to denote stop-gradient.
A frozen reference policy $\pi_{\mathrm{ref}}$ is additionally kept for target-language anchoring.
All terms are applied only to response tokens, indicated by a completion mask $m_t\in\{0,1\}$, and $N=\sum_{t=1}^{T}m_t$.
All KL divergences are computed over the full vocabulary.

\subsection{Teacher Views}
\label{sec:method_teacher_views}

Prior work attributes low-resource reasoning failures to two factors: question understanding and multilingual reasoning \citep{liu2026trit}.
To control for the first factor, both teacher views are given the English translation of the question, denoted by $x^h$.
Thus, the contrast between the two views is designed to reflect the effect of privileged reasoning information, rather than a failure to understand the low-resource question.

The \textbf{solution-conditioned view} receives the low-resource question, the English translation of the question, the English reference trace:
\[
q_t^{+}
=
\operatorname{sg}\!\left[
\pi_\theta(\cdot \mid x^\ell, x^h, s^h, y_{<t})
\right].
\]

The \textbf{ablated view} keeps the same question information and prefix, but removes the reference trace:
\[
q_t^{-}
=
\operatorname{sg}\!\left[
\pi_\theta(\cdot \mid x^\ell, x^h, y_{<t})
\right].
\]

Both $q_t^{+}$ and $q_t^{-}$ predict the next token after the same low-resource rollout prefix.
They differ only in whether the privileged reasoning trace is available.
Their distributional gap therefore probes where privileged reasoning evidence changes the model's target-language generation preference.

\begin{figure*}[!t]
  \centering
  \includegraphics[width=\textwidth]{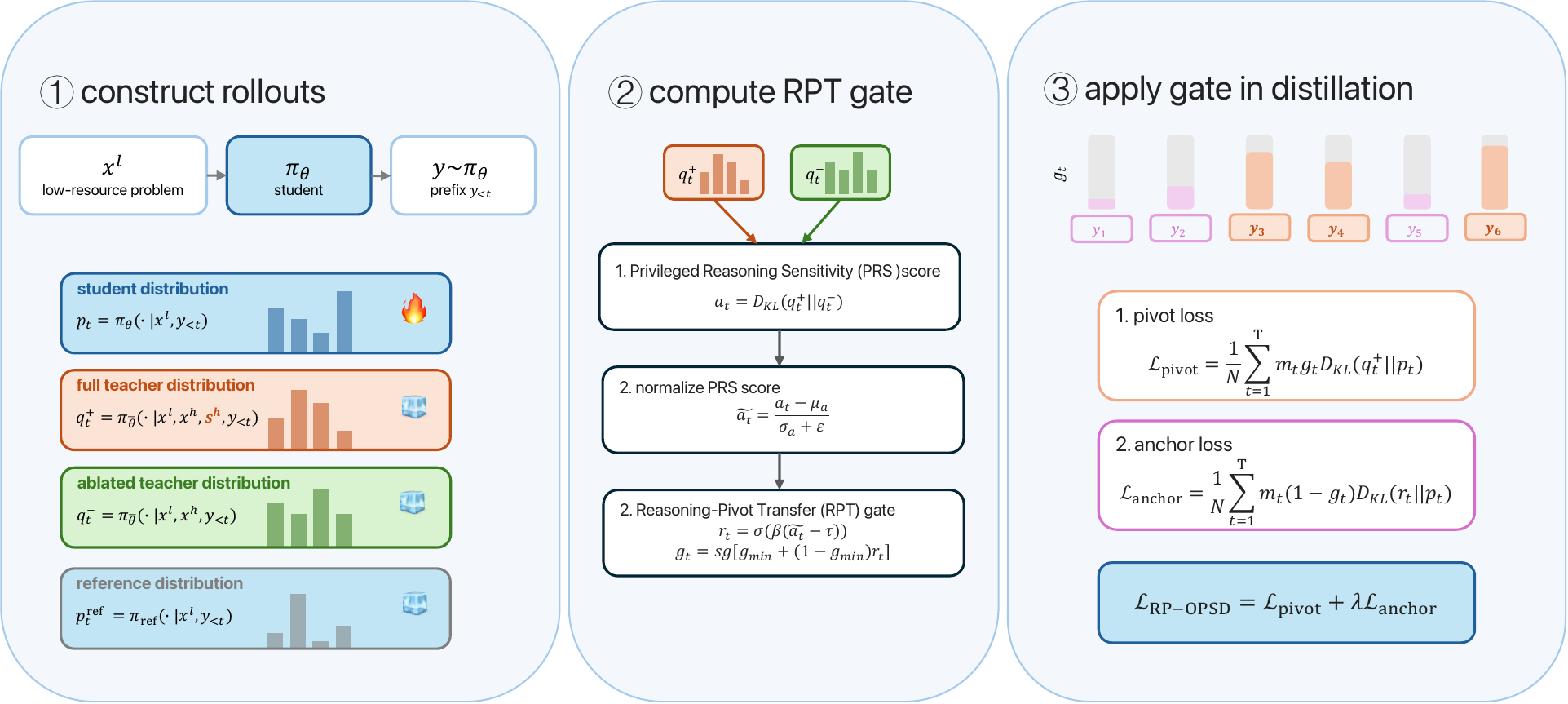}
  \caption{Overview of the RP-OPSD pipeline. RP-OPSD constructs on-policy target-language rollouts, compares solution-conditioned and ablated teacher views to compute the RPT gate, and applies the gate to route distillation toward reasoning-sensitive tokens while anchoring surface-realization tokens to the reference distribution.}
  \label{fig:pipeline}
\end{figure*}

\subsection{Reasoning-Pivot Transfer Gate}
\label{sec:method_rpt_gate}

The Reasoning-Pivot Transfer (RPT) gate is a token-level routing coefficient.
It assigns larger privileged-transfer weight to positions where the reference trace substantially changes the teacher distribution.
We denote the Privileged Reasoning Sensitivity (PRS) score by $a_t$:
\[
a_t
=
D_{\mathrm{KL}}\!\left(q_t^{+}\,\|\,q_t^{-}\right).
\]

The forward direction measures whether the ablated view can explain the probability mass emphasized by the solution-conditioned view.
A large $a_t$ indicates that the privileged trace materially changes the next-token preference at position $t$, suggesting proximity to a reasoning pivot.

Since the scale of $a_t$ varies across models, languages, and training stages, RP-OPSD normalizes it with running statistics over completion tokens:
\[
\tilde{a}_t
=
\frac{a_t-\mu_a}{\sigma_a+\epsilon}.
\]

The RPT gate is then defined as
\[
g_t
=
\operatorname{sg}\!\left[
g_{\min}
+
(1-g_{\min})
\sigma\!\left(\beta(\tilde{a}_t-\tau)\right)
\right],
\]
where $\sigma(\cdot)$ is the sigmoid function.
Here $\beta$ controls gate sharpness, $\tau$ is the threshold, and $g_{\min}$ prevents the privileged signal from vanishing entirely.
High $g_t$ routes the token toward solution-conditioned distillation; low $g_t$ treats the token as primarily governed by target-language realization or generic decoding priors.

\subsection{Routed Training Terms}
\label{sec:method_objective}

RP-OPSD uses two full-vocabulary forward-KL terms, both computed on the on-policy response.
The pivot-transfer term aligns the student with the solution-conditioned teacher view at positions selected by the RPT gate:
\[
\mathcal{L}_{\mathrm{pivot}}
=
\frac{1}{N}
\sum_{t=1}^{T}
m_t g_t
D_{\mathrm{KL}}\!\left(q_t^{+}\,\|\,p_t\right).
\]
This term concentrates privileged-context supervision on reasoning-sensitive decisions, where the English reference trace changes the target-language next-token distribution.

For language anchoring, let $r_t$ denote the stop-gradient next-token distribution of the frozen reference policy $\pi_{\mathrm{ref}}$, conditioned on the low-resource input $x^\ell$ and the rollout prefix $y_{<t}$.
The anchoring term is then
\[
\mathcal{L}_{\mathrm{anchor}}
=
\frac{1}{N}
\sum_{t=1}^{T}
m_t(1-g_t)
D_{\mathrm{KL}}\!\left(r_t\,\|\,p_t\right).
\]
This term preserves low-resource surface realization at positions where privileged reasoning evidence adds little next-token information.

The RP-OPSD training objective combines the two routed terms:
\[
\mathcal{L}_{\mathrm{RP\text{-}OPSD}}
=
\mathcal{L}_{\mathrm{pivot}}
+
\lambda\mathcal{L}_{\mathrm{anchor}}.
\]
The coefficient $\lambda$ controls the strength of reference anchoring.
All teacher, reference, score, and gate quantities are detached; gradients are taken only through the student distribution $p_t$.

Figure~\ref{fig:pipeline} summarizes the full RP-OPSD pipeline: the student first constructs on-policy target-language rollouts, the two teacher views then define the RPT gate through the PRS score, and the gate finally routes token-level supervision between pivot transfer and language anchoring.

\section{Experiment}

\begin{table*}[t]
\centering
\scriptsize
\setlength{\tabcolsep}{2pt}
\resizebox{\textwidth}{!}{%
\begin{tabular}{cl*{20}{c}}
\toprule
\multirow{2}{*}{} & \multirow{2}{*}{Method}
& \multicolumn{13}{c}{AfriMGSM \textcolor{gray}{\textit{(pass@12)}}}
& \multicolumn{7}{c}{PolyMath \textcolor{gray}{\textit{(DW-ACC)}}} \\
\cmidrule(lr){3-15}\cmidrule(lr){16-22}
& & AMH & EWE & HAU & KIN & LUG & SNA & SOT & SWA & WOL & XHO & YOR & ZUL & Avg.
& ZHO & FRA & SWA & JPN & SPA & RUS & Avg. \\
\midrule
\multirow{8}{*}{\rotatebox[origin=c]{90}{Qwen3-1.7B}} & Base
& 12.0 & 8.8 & 13.2 & 9.2 & 7.2 & 8.8 & 8.4 & 13.2 & 8.8 & 9.2 & 11.2 & 8.8 & 9.90
& 21.17 & 17.98 & 0.85 & 12.96 & 16.43 & 19.84 & 14.87 \\
\noalign{\vskip 1.4pt}
\cdashline{2-22}[0.35pt/1.6pt]
\noalign{\vskip 2.2pt}
& SFT
& 12.8 & 9.2 & 14.0 & \underline{14.4} & 11.2 & 11.2 & 12.0 & 16.8 & \underline{14.4} & \underline{15.6} & 10.0 & 13.2 & 12.90
& 20.38 & 16.97 & 1.44 & 11.95 & 15.58 & 18.77 & 14.18 \\
& GRPO
& 13.2 & 10.4 & 13.2 & 7.6 & 6.8 & 9.2 & 9.6 & 15.2 & 6.0 & 6.8 & 12.4 & 10.0 & 10.03
& 24.16 & 18.51 & 0.27 & 13.01 & 18.67 & 18.29 & 15.48 \\
\noalign{\vskip 1.4pt}
\cdashline{2-22}[0.35pt/1.6pt]
\noalign{\vskip 2.2pt}
& MAPO-DPO
& 12.0 & 6.4 & 11.6 & 7.6 & 4.8 & 8.0 & 6.0 & 13.6 & 7.6 & 6.0 & 10.0 & 9.2 & 8.57
& 23.68 & 18.13 & 0.75 & 13.60 & 17.55 & 19.47 & 15.53 \\
& M-Thinker
& 22.8 & 13.6 & 13.6 & 10.4 & 14.0 & 10.0 & 14.4 & 22.4 & 13.2 & 14.4 & 12.0 & 15.6 & 14.87
& \underline{24.80} & 19.14 & 1.12 & 14.88 & 19.47 & 20.32 & 16.62 \\
& PCS
& --- & --- & --- & --- & --- & --- & --- & 16.0 & --- & --- & --- & --- & ---
& 24.69 & 19.30 & 1.39 & \underline{15.05} & \underline{19.63} & \underline{20.37} & \underline{16.74} \\
\noalign{\vskip 1.4pt}
\cdashline{2-22}[0.35pt/1.6pt]
\noalign{\vskip 2.2pt}
& COPSD
& 23.6 & \underline{14.8} & 16.4 & \underline{14.4} & 16.0 & 11.6 & 15.2 & 26.0 & 13.6 & 15.2 & 14.4 & \textbf{19.2} & \underline{16.70}
& 23.20 & \underline{19.31} & \underline{4.00} & 13.38 & 17.71 & 18.35 & 15.99 \\
& EGRSD
& \underline{24.0} & \underline{14.8} & \underline{17.2} & 14.0 & \underline{16.4} & \underline{13.6} & \underline{16.4} & \underline{26.8} & 12.8 & 11.2 & \underline{14.8} & \underline{14.8} & 16.40
& 23.31 & \underline{19.31} & 3.25 & 14.13 & 18.72 & 19.15 & 16.31 \\
& \textbf{RP-OPSD}
& \textbf{24.4} & \textbf{15.2} & \textbf{20.0} & \textbf{16.0} & \textbf{18.4} & \textbf{19.2} & \textbf{17.6} & \textbf{29.6} & \textbf{15.2} & \textbf{17.6} & \textbf{16.4} & \textbf{19.2} & \textbf{19.07}
& \textbf{25.49} & \textbf{20.05} & \textbf{5.23} & \textbf{15.31} & \textbf{20.85} & \textbf{20.91} & \textbf{17.97} \\
\midrule
\multirow{8}{*}{\rotatebox[origin=c]{90}{Qwen3-4B}} & Base
& 26.8 & \underline{17.6} & \underline{23.6} & 17.6 & 16.8 & 14.8 & 17.2 & 44.0 & 15.2 & 17.6 & \underline{18.0} & 14.4 & 20.30
& 36.27 & 34.56 & 2.24 & \underline{29.55} & 33.71 & 34.67 & 28.50 \\
\noalign{\vskip 1.4pt}
\cdashline{2-22}[0.35pt/1.6pt]
\noalign{\vskip 2.2pt}
& SFT
& 25.2 & 12.4 & 17.6 & 15.2 & 17.6 & \underline{17.2} & 18.4 & 40.4 & 17.2 & 17.2 & 11.6 & 15.2 & 18.77
& 35.47 & 32.96 & 1.92 & 26.08 & 32.21 & 33.97 & 27.10 \\
& GRPO
& 28.8 & 15.6 & 20.8 & 17.6 & 16.0 & 14.0 & \underline{19.2} & 44.8 & 15.2 & 16.8 & 17.2 & 14.4 & 20.03
& 38.19 & 31.47 & 1.92 & \textbf{29.87} & 34.61 & 34.45 & 28.42 \\
\noalign{\vskip 1.4pt}
\cdashline{2-22}[0.35pt/1.6pt]
\noalign{\vskip 2.2pt}
& MAPO-DPO
& 27.6 & 13.6 & \underline{23.6} & \underline{18.0} & 18.0 & 15.2 & 17.6 & 47.2 & 19.2 & 17.6 & 16.4 & 15.6 & 20.80
& 37.55 & 34.51 & 7.52 & 27.52 & 33.07 & 33.12 & 28.88 \\
& M-Thinker
& \underline{34.4} & 15.2 & 17.2 & \textbf{22.8} & \underline{23.2} & \underline{20.8} & 15.6 & 50.4 & 19.2 & 18.0 & 16.4 & 17.2 & 22.53
& 38.35 & 34.51 & 8.06 & 28.49 & 35.52 & 33.49 & 29.74 \\
& PCS
& --- & --- & --- & --- & --- & --- & --- & 51.6 & --- & --- & --- & --- & ---
& \underline{38.61} & 34.72 & \underline{8.32} & 28.70 & \underline{35.57} & 34.34 & 30.04 \\
\noalign{\vskip 1.4pt}
\cdashline{2-22}[0.35pt/1.6pt]
\noalign{\vskip 2.2pt}
& COPSD
& \underline{34.4} & 15.6 & 20.8 & 16.0 & 18.8 & 17.2 & 16.4 & 51.2 & \underline{20.0} & 14.0 & 17.2 & 18.0 & 21.63
& 37.34 & 34.77 & 7.68 & 29.49 & 35.10 & 35.26 & 29.94 \\
& EGRSD
& \underline{34.4} & 17.2 & 22.0 & \underline{18.0} & 20.0 & 16.4 & 16.8 & \underline{51.8} & 19.6 & \underline{20.0} & 14.8 & \underline{21.2} & \underline{22.68}
& 37.68 & \underline{34.93} & 8.00 & 29.39 & 35.37 & \underline{35.58} & \underline{30.16} \\
& \textbf{RP-OPSD}
& \textbf{35.2} & \textbf{22.4} & \textbf{26.4} & \textbf{22.8} & \textbf{26.0} & \textbf{25.2} & \textbf{22.8} & \textbf{54.8} & \textbf{20.8} & \textbf{22.8} & \textbf{19.2} & \textbf{23.6} & \textbf{26.83}
& \textbf{39.99} & \textbf{35.04} & \textbf{12.11} & 29.44 & \textbf{37.01} & \textbf{37.65} & \textbf{31.87} \\
\bottomrule
\end{tabular}%
}
\caption{Main results on AfriMGSM and PolyMath. Averages are arithmetic means over all reported languages; the AfriMGSM average for PCS is omitted because only Swahili is available. The metric for AfriMGSM is pass@12 (\%), while the metric for PolyMath is difficulty-weighted ACC@1 (DW-ACC, \%). The best result for each model scale is bolded, while the second-best result is underlined. Since the language detection tools required by PCS do not support recognizing African languages other than Swahili, we did not test PCS's performance on these languages.}
\label{tab:main_results}
\end{table*}

\subsection{Experimental Setup}
\label{sec:setup}

\noindent\textbf{Models.}
We evaluate RP-OPSD on two model scales from the Qwen3 family: Qwen3-1.7B and Qwen3-4B \citep{yang2025qwen3technicalreport}. Results on an additional model family are provided in Appendix~\ref{app:phi4_results}.

\noindent\textbf{Languages.}
We consider 17 target languages in total. 
The low-resource setting consists of 12 African languages, while the medium- and high-resource setting consists of Chinese (ZHO), French (FRA), Japanese (JPN), Spanish (SPA) and Russia (RUS). 
For each target language, we train a separate adapted model, rather than a single model shared across target languages.

\noindent\textbf{Datasets.}
For \textit{training}, we sample 500 examples from OpenThoughts \citep{guha2025openthoughtsdatarecipesreasoning}, following COPSD ~\citep{liu2026copsd}. Each example contains an English problem statement and its target-language translation, an English step-by-step solution, and a final answer.
For \textit{evaluation}, we use AfriMGSM \citep{adelani2025irokobenchnewbenchmarkafrican} for the 12 African languages and report pass@12, following COPSD \citep{liu2026copsd}. For the five medium- and high-resource languages and Swahili, we use PolyMath \citep{wang2025polymathevaluatingmathematicalreasoning}, a relatively difficult benchmark, which contains 1,000 problems in each language and is catogrized by 4 difficulty levels. We report difficulty-weighted accuracy (DW-ACC) for PolyMath, which weights each problem by its difficulty level. Results on other datasets are provided in Appendix~\ref{app:mgsm_results}.

\noindent\textbf{Baselines.}
We compare RP-OPSD with seven baselines spanning three categories. 
As standard training methods, \emph{SFT} fine-tunes the model on the 500 OpenThoughts training instances translated into each target language by DeepSeek-V4-Flash \citep{deepseekai2026deepseekv4highlyefficientmilliontoken}, while \emph{GRPO} \citep{shao2024deepseekmathpushinglimitsmathematical} performs on-policy reinforcement learning using answer correctness as the rollout-level reward. 
As representative multilingual reasoning alignment methods, \emph{MAPO} \citep{she2024mapoadvancingmultilingualreasoning} turns translation-based alignment between target-language and English reasoning traces into a preference-optimization signal; following the original setup, we use its MAPO-DPO variant and score the alignment with NLLB-200-distilled-600M \citep{nllbteam2022languageleftbehindscaling}. \emph{M-Thinker} \citep{zhang-etal-2026-think} improves multilingual reasoning through GRPO with a language-consistency reward that enforces consistency across the input, reasoning trace, and answer, and a cross-lingual thinking-alignment reward that transfers reasoning capabilities from English to target languages. \emph{PCS} \citep{wang2026efficientmultilingualreasoningtransfer} transfers English reasoning to target languages through progressive code-switching. Due to language detection tool support issues required by PCS, we only tested PCS's performance on a limited number of languages.
As OPSD variants, \emph{COPSD} \citep{liu2026copsd} uses the model itself as a privileged teacher conditioned on the English problem and reference solution, providing dense token-level supervision for target-language rollouts, while \emph{EGRSD} \citep{ke2026respectingselfuncertaintyonpolicyselfdistillation} reweights token-level distillation according to teacher confidence.

\subsection{Main Results}
\label{sec:main_results}

Table~\ref{tab:main_results} shows that RP-OPSD consistently improves multilingual reasoning at both model scales. On AfriMGSM, RP-OPSD reaches average pass@12 scores of 19.07 and 26.83 for Qwen3-1.7B and Qwen3-4B, respectively, improving over COPSD by 2.37 and 5.20 points and over M-Thinker, the strongest fully reported non-OPSD baseline, by 4.20 and 4.30 points. RP-OPSD also outperforms EGRSD at both scales. Across individual languages, RP-OPSD matches or outperforms the conventional multilingual reasoning transfer baselines MAPO-DPO, M-Thinker, and PCS in every comparable AfriMGSM setting, suggesting that reasoning-pivot-aware distillation is more effective than full-trace alignment, language-reward-guided RL, or progressive code-switching.

{\setlength{\parskip}{0pt}
On PolyMath, RP-OPSD achieves the highest average DW-ACC at both scales, reaching 17.97 and 31.87 and improving over COPSD by 1.98 and 1.93 points, respectively. It also exceeds M-Thinker by 1.35 and 2.13 points. Although pass@12 and DW-ACC are not directly comparable, the gains over the base models are numerically more modest on PolyMath (+3.10 and +3.37 DW-ACC) than on AfriMGSM (+9.17 and +6.53 pass@12). This pattern is consistent with DW-ACC assigning greater weight to difficult problems, thereby emphasizing the cases where further improvements are hardest to obtain. Overall, RP-OPSD yields consistent gains in both low-resource and broader multilingual mathematical reasoning. Further experiments demonstrate that these gains extend across additional benchmarks, model families, and domains; see Appendices~\ref{app:mgsm_results}, \ref{app:phi4_results}, and~\ref{app:out_of_domain_generalization}.
\par}

\section{Ablation and Analysis}

\subsection{Ablation and Functional Analysis of the RPT Gate}
\label{sec:rpt_gate_analysis}

Table~\ref{tab:ablation} jointly ablates RP-OPSD and functionally evaluates the RPT gate on SWA and FRA, using Base and the full RP-OPSD objective as reference points. We construct three budget-matched hard-routing variants that apply privileged distillation to 20\% of rollout tokens, with no reference anchoring on the remaining 80\%: \emph{TG} selects the highest-gate tokens, \emph{BG} the lowest-gate tokens, and \emph{RG} a random subset. Thus, TG--RG tests whether gate-based localization matters beyond supervision quantity, while TG--BG tests whether the gate ranking correctly orders transfer utility. We additionally remove reference anchoring while retaining continuous RPT-weighted distillation to isolate the contribution of the anchoring branch.

\begin{table}[t]
\centering
\small
\setlength{\tabcolsep}{4.5pt}
\begin{tabular}{lcc}
\toprule
Configuration
& \shortstack{SWA\\pass@12 (\%)}
& \shortstack{FRA\\ACC@1 (\%)} \\
\midrule
Base & 13.2 & 69.2 \\
\cdashline{1-3}
\multicolumn{3}{l}{\textit{Hard-gate localization (20\% token budget)}} \\
\quad TG & 26.0 & \underline{74.8} \\
\quad BG & 15.6 & 73.2 \\
\quad RG & 20.8 & 74.0 \\
\cdashline{1-3}
\multicolumn{3}{l}{\textit{Objective ablation}} \\
\quad w/o reference anchoring & \underline{27.6} & 73.6 \\
\midrule
\textbf{RP-OPSD} & \textbf{29.6} & \textbf{76.8} \\
\bottomrule
\end{tabular}
\caption{Ablation and functional analysis of the RPT gate on Qwen3-1.7B. SWA and FRA are evaluated on AfriMGSM with pass@12 and MGSM with ACC@1, respectively. TG, BG, and RG select the top, bottom, and random 20\% of tokens for privileged distillation under a matched token budget; the remaining tokens receive neither privileged distillation nor reference anchoring. The objective ablation retains continuous RPT-weighted distillation but removes reference anchoring. Bold and underlined values denote the best and second-best result in each column.}
\label{tab:ablation}
\end{table}

Under the matched 20\% budget, TG reaches 26.0 on SWA and 74.8 on FRA, compared with 20.8/74.0 for RG and 15.6/73.2 for BG. The consistent ordering TG $>$ RG $>$ BG supports the interpretation that RPT scores rank where privileged reasoning supervision is most useful, rather than merely identifying tokens that benefit from arbitrary teacher exposure. The localization advantage is substantially larger on SWA, where TG exceeds RG and BG by 5.2 and 10.4 points, than on FRA, where the corresponding margins are 0.8 and 1.6 points. This asymmetry suggests that precise localization matters most when target-language reasoning is the primary bottleneck, whereas the stronger FRA model can obtain modest calibration benefits from less selective supervision. Full RP-OPSD performs best overall at 29.6/76.8, showing that hard token selection is a useful diagnostic but not a replacement for the complete objective. Removing reference anchoring lowers performance to 27.6/73.6; its larger effect on FRA ($-3.2$ versus $-2.0$ points on SWA) is consistent with anchoring being particularly important when useful target-language behavior already exists and must be protected from unnecessary privileged-teacher pressure. Together, these results support a two-way routing interpretation: the RPT ranking identifies where reasoning transfer is most valuable, while reference anchoring constrains unnecessary distribution shift elsewhere.

\subsection{What Does the RPT Gate Identify?}
\label{sec:pivot_token_categories}

To characterize high-gate positions, we analyze 256 Chinese CoTs generated by Qwen3-1.7B and select the top and bottom 20\% of tokens within each CoT as pivot and surface candidates. Corpus frequency retrieves \emph{case-shared pivots}, Gate-TF-IDF retrieves \emph{rare pivots}, and a contextual-residual score retrieves \emph{common but case-specific pivots}. Table~\ref{tab:pivot_token_examples} gives representative contexts, with ranking details in Appendix~\ref{app:gate_pivot_categories}.

\begin{table*}[!t]
\centering
\begin{CJK*}{UTF8}{gbsn}
\scriptsize
\setlength{\tabcolsep}{2.2pt}
\renewcommand{\arraystretch}{0.98}
\renewcommand{\CJKboldshift}{0.025em}
\newcommand{\pivtok}[1]{{\fontseries{bx}\selectfont #1}}
\newcommand{\rawbs}{\textbackslash}
\begin{tabular}{@{}c p{0.425\textwidth}:p{0.425\textwidth}@{}}
\toprule
& \textbf{Pivot/Surface Context} & \textbf{English Translation} \\
\midrule
    \multirow{5}{*}{\rotatebox[origin=c]{90}{\shortstack{Case-Shared\\Pivots}}}
& \ldots 符合条件，\pivtok{所以}答案是(A) 9。
& \ldots satisfies the condition, \pivtok{so} the answer is (A) 9. \\
& \ldots 这里\pivtok{可能}哪里出错了？
& \ldots What \pivtok{might} have gone wrong here? \\
& \ldots \pivtok{但}需要确认思路是否正确。
& \ldots \pivtok{but} we need to verify whether the reasoning is correct. \\
& \ldots \pivtok{或者}考虑使用几何变换。
& \ldots \pivtok{or} consider using a geometric transformation. \\
& \ldots \pivtok{因此}交点为\texttt{(16,...)}。
& \ldots \pivtok{Therefore}, the intersection point is \texttt{(16,...)}. \\
\noalign{\vskip 0.5pt}
\cdashline{1-3}[0.35pt/1.6pt]
\noalign{\vskip 1.0pt}
\multirow{5}{*}{\rotatebox[origin=c]{90}{\shortstack{Rare\\Pivots}}}
& \ldots 可能指\pivtok{梯}形的周长？
& \ldots Could it refer to the perimeter of the \pivtok{trapezoid}? \\
& \ldots 三角形B\pivtok{DE}和CDE的面积比例。
& \ldots the ratio of the areas of triangles B\pivtok{DE} and CDE. \\
& \ldots 1001是\pivtok{四位}回文数。
& \ldots 1001 is a \pivtok{four-digit} palindrome. \\
& \ldots 正多边形的\pivtok{旋转}对称性。
& \ldots the \pivtok{rotational} symmetry of a regular polygon. \\
& \ldots 第一次\pivtok{翻}转的硬币。
& \ldots the coin \pivtok{flipped} for the first time. \\
\noalign{\vskip 0.5pt}
\cdashline{1-3}[0.35pt/1.6pt]
\noalign{\vskip 1.0pt}
    \multirow{5}{*}{\rotatebox[origin=c]{90}{\shortstack{Common but\\Case-Specific\\Pivots}}}
& \texttt{(5\rawbs cos\rawbs\pivtok{theta}, 5\rawbs sin\rawbs theta)}
& \texttt{(5\rawbs cos\rawbs\pivtok{theta}, 5\rawbs sin\rawbs theta)} \\
& \ldots 结合三角\pivtok{形}的边角关系。
& \ldots combine this with the side--angle relations of the \pivtok{triangle}. \\
& \texttt{(1-\rawbs\pivtok{sqrt}\{\rawbs cos x\})(1+\rawbs sqrt\{\rawbs cos x\})}
& \texttt{(1-\rawbs\pivtok{sqrt}\{\rawbs cos x\})(1+\rawbs sqrt\{\rawbs cos x\})} \\
& \ldots 在中文\pivtok{里}，可能用21:10。
& \ldots \pivtok{in} Chinese, 21:10 might be used. \\
& \texttt{\rawbs sqrt[\pivtok{\{n}...]}
& \texttt{\rawbs sqrt[\pivtok{\{n}...]} \\
\noalign{\vskip 0.5pt}
\cdashline{1-3}[0.35pt/1.6pt]
\noalign{\vskip 1.0pt}
\multirow{5}{*}{\rotatebox[origin=c]{90}{\shortstack{Frequent\\Surface\\Tokens}}}
& \texttt{y1=e\textasciicircum\{\pivtok{x}\}}，所以其导数\ldots
& \texttt{y1=e\textasciicircum\{\pivtok{x}\}}, so its derivative \ldots \\
& \ldots \texttt{5a+3\pivtok{b}}和\texttt{13a+8b}。
& \ldots \texttt{5a+3\pivtok{b}} and \texttt{13a+8b}. \\
& \texttt{a\textasciicircum2=(a+x)(\pivtok{a}+x/3)}
& \texttt{a\textasciicircum2=(a+x)(\pivtok{a}+x/3)} \\
& \ldots \texttt{N-1}是一个素\pivtok{数}。
& \ldots \texttt{N-1} is a prime \pivtok{number}. \\
& \texttt{\rawbs\pivtok{frac}\{4ax\}\{3\}}
& \texttt{\rawbs\pivtok{frac}\{4ax\}\{3\}} \\
\bottomrule
\end{tabular}
\end{CJK*}
\caption{Gate-derived pivot and surface contexts with English translations; bold marks the ranked tokenizer unit. Rare and common but case-specific pivots use Gate-TF-IDF and contextual-residual rankings.}
\label{tab:pivot_token_examples}
\end{table*}

\begin{CJK*}{UTF8}{gbsn}
The rankings reveal two functional types of reasoning pivots. The first is \emph{reasoning-control pivots}, including ``所以'' (so), ``因此'' (therefore), and ``但'' (but). They signal conclusions, consequences, and revisions that organize the reasoning trajectory. This pattern agrees with prior findings on information-rich transitional tokens and influential planning or uncertainty-management steps \citep{qian2025demystifyingreasoningdynamics,bogdan2025thoughtanchorsllmreasoning}. Further analysis is provided in Appendix~\ref{app:thought_anchor_alignment}.

The second type is \emph{problem-conditioned state-update pivots}, captured by the rare and common but case-specific rankings. Examples include ``梯'' in ``梯形'' (trapezoid), ``线段 \texttt{DE}'' (segment \texttt{DE}), and ``平方根 \texttt{sqrt}'' (square-root operator). They encode the problem-specific concepts, variables, and operations that advance the current solution state.
\end{CJK*}

Frequent surface tokens include \texttt{x}, \texttt{b}, and \texttt{frac}. They mainly realize variables and formula syntax after the local reasoning step has been determined. This contrast shows that the RPT gate identifies transferable reasoning decisions while preserving routine target-language and symbolic realization, supporting its use for selective privileged distillation.

\subsection{RP-OPSD Extends Reasoning beyond Base-Model Reachability}
\label{sec:beyond_base_reachability}

This section discusses the theoretical advantages and experimental support of RP-OPSD over other methods in the baseline for improving model reasoning capabilities. RP-OPSD requires no translated or generated target-language solution: only the question is translated, while the reference solution remains in English. M-Thinker and PCS instead introduce target-language rationale tokens during cold-start SFT \citep{zhang-etal-2026-think,wang2026efficientmultilingualreasoningtransfer}; these traces mainly bootstrap target-language generation, while the larger reasoning gains emerge during subsequent RL. RP-OPSD keeps the external English solution active in its core objective, with the solution-conditioned teacher supplying a full next-token distribution along the student's on-policy target-language rollout. M-Thinker's CTA must first sample a correct English trace, while PCS relies on outcome-level rollout rewards. When the current policy cannot discover a correct trajectory, neither RL objective provides a direct solution-conditioned token target, whereas RP-OPSD remains supervised by the supplied English solution.

\begin{figure}[!t]
  \centering
  \includegraphics[width=\linewidth]{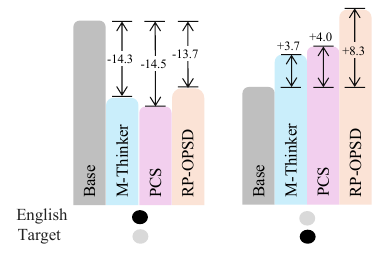}
  \caption{Cross-lingual solution reachability on PolyMath before and after training, averaged over six target languages. Filled circles (\textcolor{black}{$\bullet$}) indicate correct answers and hollow circles (\textcolor{gray}{$\bullet$}) incorrect answers; accordingly, the upper group compares the number of English-correct but target-incorrect problems before and after training, whereas the lower group compares the number of English-incorrect but target-correct problems. Signed annotations report mean changes from the untrained Qwen3-1.7B base model; bar lengths are schematic and do not represent absolute proportions.}
  \label{fig:beyond_base_reachability}
\end{figure}

Figure~\ref{fig:beyond_base_reachability} separates gains that \emph{transfer} an English-reachable solution from gains that extend beyond English reachability. M-Thinker and PCS reduce, on average, the English-correct but target-incorrect set by 14.3 and 14.5 problems, respectively, while enlarging the English-incorrect but target-correct set by only 3.7 and 4.0. RP-OPSD achieves a comparable 13.7-problem reduction in the former set but an 8.3-problem increase in the latter. It therefore retains 94.5\% of the strongest baseline's English-reachable transfer gain while more than doubling its beyond-English-reachability gain ($2.08\times$ over PCS). The improvement is consequently not explained only by re-expressing reasoning already accessible in English: dense access to an external English solution expands the target-language solution set into a region that outcome-driven RL rarely reaches from the base policy.

\subsection{Language Consistency under Reasoning Transfer}
\label{sec:language_consistency}

\begin{figure}[!t]
  \centering
  \includegraphics[width=\linewidth]{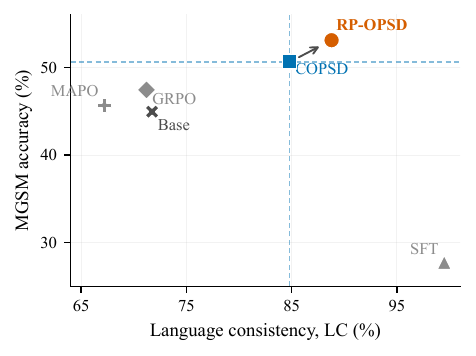}
  \caption{Language consistency (LC) versus MGSM accuracy on the focused AMH, SWA, ZH, KR, and TH subset. Dashed lines mark COPSD, and the arrow shows the shift from COPSD to RP-OPSD.}
  \label{fig:language_consistency_focused}
\end{figure}

\begin{figure*}[!t]
  \centering
  \includegraphics[width=\textwidth]{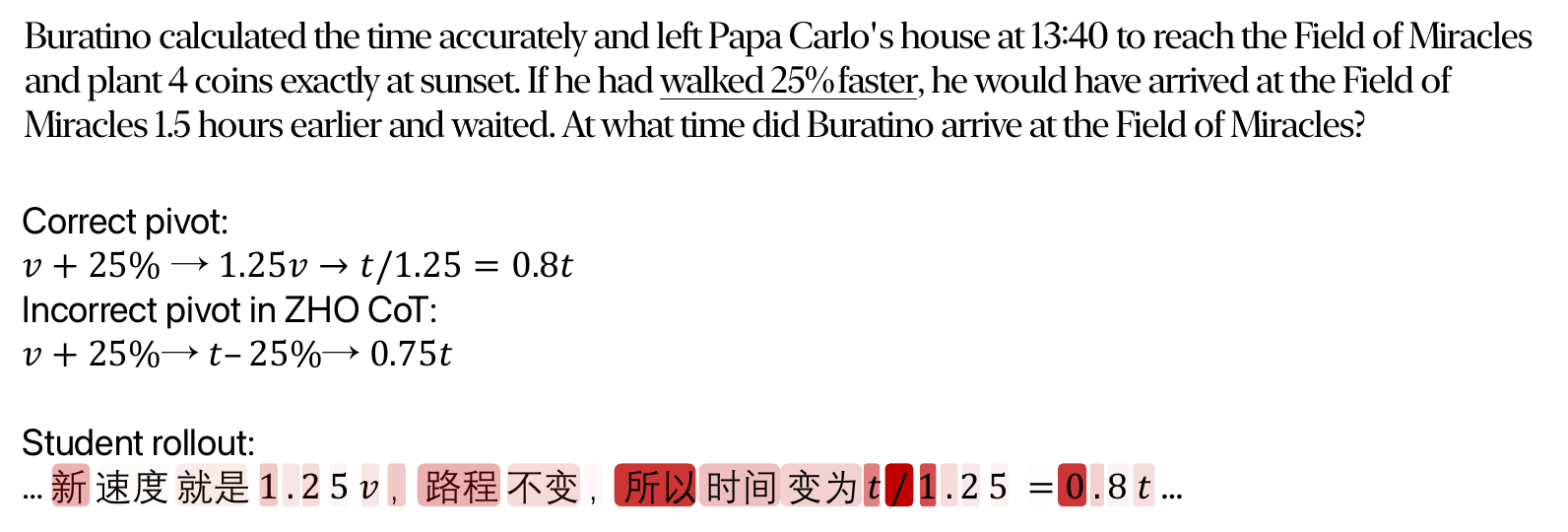}
  \caption{Case study of RPT-gate localization. The upper lines contrast the correct time-rescaling pivot with an incorrect Chinese CoT pivot. In the student rollout, each token is colored by its normalized RPT-gate value, with warmer colors indicating larger transfer weights.}
  \label{fig:case_study}
\end{figure*}

A natural concern is that stronger cross-lingual reasoning transfer may simply make the model reason in a high-resource language and recover the target language only near the final answer. Figure~\ref{fig:language_consistency_focused} does not support this failure mode. Compared with COPSD, RP-OPSD moves to the upper-right of the accuracy--LC plane, improving MGSM accuracy from 50.6 to 53.1 while increasing LC from 84.8 to 88.8 on the focused subset. This movement is more informative than LC alone: SFT attains almost perfect LC but remains far below the reasoning methods in accuracy, whereas GRPO and MAPO improve accuracy by drifting into a lower-LC region. RP-OPSD is therefore not merely preserving surface language, nor is it buying accuracy through target-language erosion.

If RP-OPSD relied on English or another high-resource language as a hidden scratchpad, its point should move toward the GRPO/MAPO regime or show higher leakage; instead, its English leakage is slightly lower than COPSD on this subset (0.87\% vs. 1.04\%). On the broader seven-language comparable set, RP-OPSD also matches COPSD's LC (85.0) while improving accuracy by 2.8 points. The consistent pattern suggests that the gate changes the locus of transfer: privileged supervision is routed to tokens that alter the reasoning state, while reference anchoring preserves target-language realization elsewhere. The gain is thus a better partition between transferable reasoning and language expression, not a tradeoff between the two.

\subsection{Case Study}
\label{sec:case_study}

Figure~\ref{fig:case_study} analyzes a Chinese training rollout from Qwen3-1.7B on a time-rescaling problem. The vanilla model answers correctly in English, but its Chinese CoT interprets ``speed increases by 25\%'' as ``time decreases by 25\%,'' yielding the wrong conclusion $0.75t$. The correct transition is not a surface translation choice: increasing speed from $v$ to $1.25v$ makes the required time $t/1.25=0.8t$. Thus, the local mapping $v\mapsto1.25v \Rightarrow t\mapsto t/1.25=0.8t$ is the reasoning pivot in this example, because it changes the latent solution state rather than merely restating a quantity from the problem.

\begin{CJK*}{UTF8}{gbsn}
The RPT heatmap localizes privileged transfer around this transition. Most lexical tokens, punctuation marks, and inherited quantities remain near the background gate level, while the gate rises sharply at the division token in $t/1.25$, where the rollout commits to reciprocal time scaling; this position receives a nearly saturated gate above $0.99$. The leading digit in $0.8t$ also receives a high gate, which is less aligned with our expectation. Inspecting the teacher distributions suggests that this high value is mainly a formatting effect: the solution-conditioned teacher has a stronger preference for a fractional continuation such as \texttt{\textbackslash frac\{t\}\{1.25\}}, creating a large PRS gap at the decimal-form position. Meanwhile, “所以” (so) as a reasoning-control pivot also receives a high RPT gate, consistent with its role in signaling the continuation of the reasoning tragectory, as discussed in Section~\ref{sec:pivot_token_categories}.
\end{CJK*}

This case clarifies why RP-OPSD should not treat numbers or formulas as pivots by default. Tokens such as $25\%$ or $1.25$ can be inherited from the question or from a deterministic rewrite without requiring new reasoning. What requires transfer is the operation that binds the speed increase to reciprocal time scaling. By concentrating full-distribution distillation near that operation while leaving most target-language realization anchored, RP-OPSD avoids diluting the privileged reasoning signal over fluent but reasoning-insensitive tokens. The qualitative evidence therefore complements the ablations: the benefit of the gate comes from preserving the sparsity and polarity of cross-lingual reasoning transfer, not from assigning uniformly larger weight to mathematical-looking tokens.

\FloatBarrier

\section{Related Work}

\noindent\textbf{Multilingual Reasoning Transfer.}
Recent work on multilingual reasoning transfer has progressed from inference-time cross-lingual prompting to training-time control of reasoning language.
Cross-lingual Prompting aligns reasoning across languages and aggregates multilingual reasoning paths to improve zero-shot CoT ~\citep{qin2023crosslingualpromptingimprovingzeroshot}, while Question Translation Training performs question-level alignment by translating non-English problems into English before solution generation ~\citep{zhu2024questiontranslationtrainingbetter}.
MAPO further aligns non-dominant-language reasoning traces with English via translation-based preference optimization ~\citep{she2024mapoadvancingmultilingualreasoning}, and TRIT jointly trains translation and reasoning to enhance multilingual understanding and target-language generation without extra supervision ~\citep{liu2026trit}.
Complementary work studies structured prompting, English-pivoted reasoning, RL-based alignment, and how reasoning language shapes intermediate reasoning ~\citep{ranaldi2024empoweringmultistepreasoninglanguages, tran-etal-2025-disentangling, tam2025languagemattersmultilingualinput, zhang-etal-2026-think, zhang2026doesalignmentenhancellms}.
In contrast, our work shifts from aligning full reasoning traces to identifying where transfer should occur, using privileged-context distributional shifts to selectively transfer reasoning-critical signals while preserving target-language realization.

\noindent\textbf{On-Policy Self-Distillation.}
On-policy distillation reduces train--test mismatch by training models on their own rollouts ~\citep{agarwal2024onpolicydistillationlanguagemodels}.
OPSD uses a single model as both student and privileged-context teacher, providing dense token-level supervision without a separate teacher ~\citep{zhao2026selfdistilledreasoneronpolicyselfdistillation}.
Subsequent variants incorporate feedback-conditioned predictions, context internalization, cross-lingual privileged context, reflection-localized correction, and efficiency or stability enhancements ~\citep{hübotter2026reinforcementlearningselfdistillation, ye2026onpolicycontextdistillationlanguage, liu2026copsd, zhao2026rosdreflectiveonpolicyselfdistillation, zhang2026fasteffectiveonpolicydistillation, jang2026stableonpolicydistillationadaptive}.
RP-OPSD instead contrasts privileged and ablated teacher views to route full-distribution supervision toward reasoning-sensitive positions while anchoring target-language realization elsewhere.

\section{Conclusion}

We propose RP-OPSD, which contrasts teacher views to route privileged distillation toward reasoning-sensitive positions and reference anchoring elsewhere. 
Across 17 languages and two mathematical-reasoning benchmarks, RP-OPSD outperforms strong multilingual reasoning baselines and OPSD variants. 
High-gate positions exhibit recurring reasoning-control transitions and problem-conditioned state updates, whereas low-gate positions largely reflect routine target-language and symbolic realization. 
More broadly, RP-OPSD points toward multilingual reasoners that share computational structure across languages while retaining natural, language-specific forms of expression.

\bibliography{ref}

\appendix

\section{RPT Gates Co-localize with Thought Anchors}
\label{app:thought_anchor_alignment}

\begin{figure*}[!t]
  \centering
  \includegraphics[width=\textwidth]{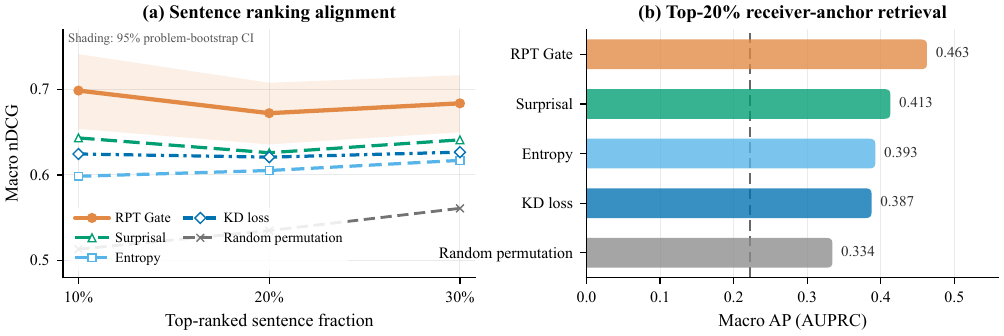}
  \caption{Sentence level alignment with thought anchors defined by receiver scores on AMH. \textbf{Left:} macro nDCG measures agreement with the graded receiver importance ranking at the top 10\%, 20\%, and 30\% of sentences; RPT shading shows the 95\% problem bootstrap confidence interval. \textbf{Right:} macro AUPRC measures retrieval of the top 20\% receiver anchors; the dashed line marks their prevalence. Baselines are entropy (student uncertainty), surprisal (negative log probability of the sampled token), KD loss (KL from the privileged teacher to the student), and random permutation (chance ranking).}
  \label{fig:thought_anchor_alignment}
\end{figure*}

Section~\ref{sec:rpt_gate_analysis} shows that privileged distillation is most effective at high RPT-gate positions. A complementary question concerns what makes these positions valuable. Intuitively, a reasoning pivot should exert a sustained influence on the following reasoning. A possible signature of this influence is that later steps repeatedly draw on the information introduced at that point.

This perspective motivates a proxy based on downstream information reuse.

Following \citet{bogdan2025thoughtanchorsllmreasoning}, we use the \emph{receiver score} as this proxy. The receiver score quantifies how strongly information from a reasoning sentence is carried forward through specialized receiver heads and reused in future reasoning steps. Sentences with high receiver scores are termed \emph{thought anchors} because they serve as persistent information sources for subsequent computation. The sentence-level score is computed solely from ordinary student attention and excludes both privileged teacher contexts and the distillation loss, providing an independent graded relevance signal for reasoning sentences. We also compare against entropy, surprisal, and KD loss, with random permutation as a reference.

Across both panels of Figure~\ref{fig:thought_anchor_alignment}, RPT scores above every baseline. Its macro nDCG is 0.699, 0.672, and 0.684 at the three cutoffs, with all controls lower at each cutoff, indicating agreement with the graded receiver importance ranking. RPT also reaches 0.463 AUPRC, exceeding surprisal (0.413), entropy (0.393), KD loss (0.387), and random permutation (0.334). The retrieval advantage shows that the most important anchors are especially concentrated among high gate sentences. The consistent margins over entropy, surprisal, and KD loss distinguish RPT from a generic uncertainty or high loss detector. Because receiver scores arise from independent ordinary student attention, the co-localization provides convergent mechanistic evidence that RPT selects states with internal downstream influence. Together with the functional analysis, the evidence identifies a distinctive intersection: privileged context changes the local prediction at states that later reasoning reuses, and supervising those states yields the greatest transfer benefit. This intersection explains their leverage as transfer points and characterizes a reasoning pivot as an information bottleneck whose state propagates through subsequent computation.

\section{Why the RPT Gate Uses Matched Teacher Views}
\label{app:matched_teacher_views}

The RPT gate is intended to locate positions where the privileged reference trace changes the model's reasoning state, rather than positions where the teacher and student merely disagree. To test whether the ablated teacher view is necessary for this distinction, we replace the Privileged Reasoning Sensitivity score
\[
a_t
=
D_{\mathrm{KL}}\!\left(q_t^{+}\,\|\,q_t^{-}\right)
\]
with a teacher--student alternative,
\[
a_t^{\mathrm{TS}}
=
D_{\mathrm{KL}}\!\left(q_t^{+}\,\|\,p_t\right).
\]
This alternative removes $q_t^{-}$ entirely while retaining the same on-policy rollouts, gate normalization and activation, routed distillation objective, reference anchoring, and 100-step training schedule. We evaluate Qwen3-1.7B at checkpoint 75 for ZHO and checkpoint 35 for SWA. For a controlled comparison, each row in Table~\ref{tab:gate_score_source_ablation} uses matched checkpoint and decoding settings across the two gate scores; in particular, both SWA results use tensor-parallel size one.

Table~\ref{tab:gate_score_source_ablation} shows a consistent degradation when the student replaces the ablated teacher in the gate score. On ZHO, the PolyMath difficulty-weighted accuracy decreases from 25.49 to 23.46. On SWA, AfriMGSM pass@12 decreases from 29.6\% to 27.2\%. These are controlled single-run comparisons, so we interpret the common direction of the changes rather than claim statistical significance.

The main difference between the scores is the quantity they isolate. The two teacher views share the low-resource question, its English translation, and the same target-language rollout prefix; they differ only in access to the English reference trace. Consequently, $D_{\mathrm{KL}}(q_t^{+}\|q_t^{-})$ measures the incremental effect of privileged reasoning evidence under a matched information state. In contrast, $p_t$ is conditioned only on the low-resource question and rollout prefix. The gap $D_{\mathrm{KL}}(q_t^{+}\|p_t)$ therefore also reflects question-understanding differences, target-language realization, teacher--student calibration, and the student's current adaptation state.

\begin{table}[!t]
\centering
\small
\setlength{\tabcolsep}{3.5pt}
\resizebox{\columnwidth}{!}{%
\begin{tabular}{llrrr}
\toprule
Language & Metric $\uparrow$ & $q_t^{+}\|q_t^{-}$ & $q_t^{+}\|p_t$ & $\Delta$ \\
\midrule
ZHO & PolyMath DW-ACC & \textbf{25.49} & 23.46 & $-2.03$ \\
SWA & AfriMGSM Pass@12 (\%) & \textbf{29.6} & 27.2 & $-2.4$ \\
\bottomrule
\end{tabular}%
}
\caption{Gate-score ablation on Qwen3-1.7B. The matched-view score is the original RP-OPSD gate. The teacher--student alternative removes $q_t^{-}$ and changes only the RPT score. $\Delta$ is the teacher--student result minus the matched-view result.}
\label{tab:gate_score_source_ablation}
\end{table}

A large value can identify a difficult lexical or formatting choice without identifying a reasoning pivot.

Replacing $q_t^{-}$ with $p_t$ also couples routing to the residual being optimized. Let $d_t=D_{\mathrm{KL}}(q_t^{+}\|p_t)$. Under the alternative score, the pivot-transfer contribution is proportional to $g(d_t)d_t$: tokens with large distillation residuals receive both a larger loss and a larger gate. This behavior amounts to residual-based hard-token reweighting, which can overemphasize persistent surface-form or calibration mismatches, causing error amplification. At the same time, a high gate reduces the complementary reference-anchoring weight $1-g_t$. A false positive therefore has two effects: it increases privileged-teacher pressure at a non-pivot token and removes the anchoring intended to preserve target-language realization. The original score separates \emph{where privileged evidence matters}, measured by $q_t^{+}$ versus $q_t^{-}$, from \emph{how much student correction remains}, measured by $q_t^{+}$ versus $p_t$.

This interpretation is consistent with the independent thought-anchor analysis in Appendix~\ref{app:thought_anchor_alignment}. The matched-view RPT score aligns more strongly with receiver-defined anchors than the KD-loss baseline, reaching 0.463 versus 0.387 macro AUPRC. Since the KD-loss baseline closely corresponds to teacher--student disagreement, this gap suggests that the matched contrast better localizes states with downstream reasoning influence. Thus, $q_t^{-}$ is not an additional supervision target; it acts as a matched counterfactual baseline that removes shared contextual and surface effects from the routing signal.

\section{Additional Results on MGSM and Related Benchmarks}
\label{app:mgsm_results}

\begin{table}[!t]
\centering
\small
\setlength{\tabcolsep}{4pt}
\resizebox{\columnwidth}{!}{%
\begin{tabular}{l*{5}{c}}
\toprule
Method & ZHO & FRA & POR & THA & KOR \\
\midrule
Base & 76.4 & 69.2 & \underline{76.0} & 54.8 & 68.4 \\
SFT & 42.8 & 35.6 & 42.0 & 35.2 & 31.2 \\
GRPO & \underline{77.6} & \underline{72.8} & 71.2 & \textbf{61.2} & \underline{71.2} \\
MAPO-DPO & 74.0 & 68.8 & 75.2 & \underline{59.6} & 69.2 \\
COPSD & 76.8 & 71.6 & 74.8 & 58.8 & 68.0 \\
\textbf{RP-OPSD} & \textbf{80.4} & \textbf{76.8} & \textbf{79.6} & \textbf{61.2} & \textbf{72.4} \\
\bottomrule
\end{tabular}%
}
\caption{Additional Qwen3-1.7B accuracy (\%) on medium- and high-resource languages. ZHO, FRA, and THA use MGSM; POR uses P-MMEval; KOR uses Global-MGSM. Bold and underlined values denote the best and second-best results within each language, respectively.}
\label{tab:additional_results}
\end{table}

Table~\ref{tab:additional_results} reports additional Qwen3-1.7B results for five medium- and high-resource languages. We move these results out of the main comparison because MGSM ~\citep{shi2022languagemodelsmultilingualchainofthought} and related benchmarks are comparatively easy at this model scale. The base model already attains 54.8--76.4\% accuracy across these evaluations, leaving less headroom and making them less diagnostic than AfriMGSM and PolyMath. We nevertheless retain the results for completeness and report only the 1.7B scale, for which all baseline results are available.

RP-OPSD achieves the highest accuracy on ZHO, FRA, POR, and KOR, and ties GRPO on THA. Averaged across the five languages, RP-OPSD reaches 74.1\% accuracy, compared with 70.8\% for GRPO, the strongest baseline by average. We therefore treat these results as additional evidence consistent with the main comparison rather than as a separate central claim.

\section{Generalization to Phi-4-mini-reasoning}
\label{app:phi4_results}

\begin{table}[!t]
\centering
\footnotesize
\setlength{\tabcolsep}{4pt}
\resizebox{\linewidth}{!}{%
\begin{tabular}{llcc}
\toprule
Language & Method & MGSM ACC@1 $\uparrow$ & PolyMath DW-ACC $\uparrow$ \\
\midrule
\multirow{2}{*}{ZHO}
& Base & 83.60 & 18.24 \\
& \textbf{RP-OPSD} & \textbf{85.20} & \textbf{24.53} \\
\midrule
\multirow{2}{*}{SWA}
& Base & 28.80 & 15.20 \\
& \textbf{RP-OPSD} & \textbf{32.00} & \textbf{16.85} \\
\bottomrule
\end{tabular}
}
\caption{Cross-family generalization on Phi-4-mini-reasoning. The MGSM column reports MGSM for ZHO and AfriMGSM for SWA. MGSM-family results are ACC@1 (\%), while PolyMath results are difficulty-weighted ACC@1 (DW-ACC, \%).}
\label{tab:phi4_results}
\end{table}

\begin{table*}[!t]
\begin{center}
\centering
\scriptsize
\setlength{\tabcolsep}{2.5pt}
\begin{tabular}{l*{16}{c}}
\toprule
\multirow{2}{*}{Method}
& \multicolumn{8}{c}{RUS}
& \multicolumn{8}{c}{SPA} \\
\cmidrule(lr){2-9}\cmidrule(lr){10-17}
& Phys. & Chem. & Bio. & CS & Eng. & Econ. & Bus. & Overall
& Phys. & Chem. & Bio. & CS & Eng. & Econ. & Bus. & Overall \\
\midrule
Base & 32.87 & 30.74 & 48.95 & 37.80 & 23.12 & 39.34 & 36.88 & 34.55 & 36.03 & 35.60 & \textbf{40.45} & 42.20 & 28.17 & 40.88 & 38.40 & 36.61\\
\textbf{RP-OPSD} & 32.87 & \textbf{31.36} & \textbf{50.48} & \textbf{38.53} & \textbf{23.84} & \textbf{40.29} & \textbf{37.89} & \textbf{36.47} & \textbf{36.88} & \textbf{36.04} & 40.17 & \textbf{44.40} & \textbf{29.51} & \textbf{41.47} & \textbf{38.78} & \textbf{38.18} \\
\bottomrule
\end{tabular}
\caption{Out-of-domain performance on the non-mathematical domains of MMLU-ProX in Russian (RUS) and Spanish (SPA). Base denotes Qwen3-1.7B, while RP-OPSD denotes the corresponding language-specific model trained only on the same OpenThoughts mathematics problems used in the main experiments. Columns report physics, chemistry, biology, computer science (CS), engineering, economics, business, and overall performance.}
\label{tab:out_of_domain_generalization}
\end{center}

\end{table*}

To examine whether the effectiveness of RP-OPSD depends on the Qwen
backbones used in our main experiments, we additionally evaluate the method
on Phi-4-mini-reasoning, a model from a different family ~\citep{xu2025phi4minireasoningexploringlimitssmall}. We compare each
language-specific RP-OPSD model with its corresponding unadapted
Phi-4-mini-reasoning model.

As shown in Table~\ref{tab:phi4_results}, RP-OPSD improves both MGSM-family
answer accuracy and PolyMath DW-ACC over the corresponding base model for
both ZHO and SWA. The gains therefore hold across two target languages and
two complementary mathematical-reasoning evaluations on a non-Qwen
backbone. This consistent cross-family pattern provides evidence that the
benefits of RP-OPSD arise from its reasoning-transfer mechanism rather than
from behavior specific to the Qwen model family.

\section{Generalization beyond Mathematical Reasoning}
\label{app:out_of_domain_generalization}

Table~\ref{tab:out_of_domain_generalization} evaluates whether the multilingual reasoning transfer induced by RP-OPSD generalizes beyond the training domain. Both language-specific models are trained exclusively on the same OpenThoughts mathematics problems used in the main experiments, without any supervision from MMLU-ProX~\citep{xuan-etal-2025-mmlu}, yet improve over the Qwen3-1.7B base model in most non-mathematical domains and in overall performance for both Russian and Spanish. These out-of-domain gains indicate that RP-OPSD transfers general reasoning ability across languages rather than merely fitting the mathematical training distribution.

\section{Sensitivity to the Reference-Anchoring Coefficient}
\label{app:lambda_sweep}

\begin{table}[!t]
\centering
\footnotesize
\setlength{\tabcolsep}{4pt}
\begin{tabular}{cc}
\toprule
\shortstack{Anchoring coefficient\\$\lambda$}
& \shortstack{SWA\\pass@12 (\%)} \\
\midrule
0.0 & 27.6 \\
\textbf{0.2} & \textbf{29.6} \\
0.5 & 24.8 \\
0.8 & 24.4 \\
\bottomrule
\end{tabular}
\caption{Sensitivity to the reference-anchoring coefficient on AfriMGSM
SWA with Qwen3-1.7B. The $\lambda=0$ row removes reference anchoring, while
$\lambda=0.2$ is the setting used by RP-OPSD.}
\label{tab:lambda_sweep}
\end{table}

\begin{figure*}[!t]
\centering

\begin{minipage}[t]{0.48\textwidth}
\centering
\fbox{
\begin{minipage}[t]{0.94\linewidth}
\footnotesize
\textbf{English reference translation}\par\vspace{0.5em}
\textbf{Question}: \textit{[problem\_target]}\par\vspace{0.5em}
\textbf{English translation of the question}: \textit{[problem\_english]}\par\vspace{0.5em}
\textit{Please think step by step in Swahili, and place your final answer inside \texttt{\textbackslash boxed\{\}}.}
\end{minipage}
}
\end{minipage}
\hfill
\begin{minipage}[t]{0.48\textwidth}
\centering
\fbox{
\begin{minipage}[t]{0.94\linewidth}
\footnotesize
\textbf{Ablated-teacher prompt (Swahili; SWA)}\par\vspace{0.5em}
\textbf{Swali}: \textit{[problem\_target]}\par\vspace{0.5em}
\textbf{Tafsiri ya Kiingereza ya swali}: \textit{[problem\_english]}\par\vspace{0.5em}
\textit{Tafadhali fikiri hatua kwa hatua kwa Kiswahili, na uweke jibu lako la mwisho ndani ya \texttt{\textbackslash boxed\{\}}.}
\end{minipage}
}
\end{minipage}

\caption{
Ablated-teacher prompt.
The left panel provides an English reference translation, with each prompt component shown on a separate line; the right panel shows the instantiated Swahili (SWA) version.
Unlike the full teacher, the ablated teacher does not receive the English reference solution.
}
\label{fig:ablated_teacher_prompt}
\end{figure*}

We study the effect of the reference-anchoring coefficient $\lambda$ on
Qwen3-1.7B for SWA. Table~\ref{tab:lambda_sweep} combines the no-anchoring
ablation ($\lambda=0$), the full-model result ($\lambda=0.2$), and additional
runs with stronger anchoring. The values for $\lambda=0$ and $\lambda=0.2$
match the corresponding ablation and main results in the paper; the
$\lambda=0.5$ and $0.8$ rows are additional sweep runs evaluated at
checkpoint 100.

A moderate amount of reference anchoring is beneficial: setting
$\lambda=0.2$ improves pass@12 by 2.0 points over removing the anchoring
term. Increasing $\lambda$ to 0.5 or 0.8 instead reduces pass@12 by 4.8 and
5.2 points, respectively. This suggests that overly strong anchoring can
constrain the privileged-teacher signal at reasoning pivots. We therefore
use $\lambda=0.2$ in the main experiments.

\section{Gate-Based Pivot and Surface Categories}
\label{app:gate_pivot_categories}

This section describes how we derive the three pivot categories and their
surface-token counterpart in Table~\ref{tab:pivot_token_examples} from a
collection $\mathcal{C}$ of CoTs with token-level gates.

\paragraph{Case-shared pivots and frequent surface tokens.}
For each CoT $c\in\mathcal{C}$, we regard the tokens with RPT gates in the top
20\% as \emph{pivot candidates} and those in the bottom 20\% as \emph{surface
tokens}.  Let $x_{c,i}$ denote token $i$ and $p_{c,i}\in[0,1]$ its within-CoT
RPT-gate percentile.  We pool the pivot candidates and surface tokens from all
CoTs in $\mathcal{C}$, count the occurrences of each token type in the two
groups, and report the five most frequent token types in each group. The
high-gate frequency ranking yields pivots shared across many cases.

\paragraph{Rare pivots.}
Inspired by TF-IDF, rare pivots identify tokens that receive high relative
gates within a particular CoT but occur in few CoTs overall.  Let
$N=|\mathcal{C}|$, let $\operatorname{df}(t)$ be the number of CoTs in
$\mathcal{C}$ containing token type $t$, and set $q=0.2$.  We define
\begin{equation}
  \begin{aligned}
    w_{c,i} &= \left[\frac{p_{c,i}-(1-q)}{q}\right]_{+}, \\
    \operatorname{idf}(t) &=
    \log\frac{N+1}{\operatorname{df}(t)+1}.
  \end{aligned}
\end{equation}
For each token--CoT pair, the gate-weighted TF-IDF score is
\begin{equation}
  S_{\mathrm{rare}}(t,c)=
  \log\!\left(1+\sum_{i:x_{c,i}=t}w_{c,i}\right)
  \operatorname{idf}(t),
\end{equation}
which favors locally high-gate tokens while discounting token types shared by
many CoTs.

\paragraph{Common but case-specific pivots.}
Common but case-specific pivots identify occurrences whose gate is unusually
high in the current CoT relative to the same token type in other CoTs. We
define the
cross-CoT baseline
\begin{equation}
  \mu_{-c}(t)=
  \frac{\sum_{\substack{c'\in\mathcal{C}\\c'\ne c}}\sum_j
  \mathbf{1}[x_{c',j}=t]p_{c',j}}
  {\sum_{\substack{c'\in\mathcal{C}\\c'\ne c}}\sum_j
  \mathbf{1}[x_{c',j}=t]}
\end{equation}
and rank high-gate occurrences by the positive contextual residual
\begin{equation}
  S_{\mathrm{case}}(c,i)=
  w_{c,i}\left[p_{c,i}-\mu_{-c}(x_{c,i})\right]_{+}.
\end{equation}
Thus, a recurring token can be selected when it is ordinary elsewhere but
becomes locally salient for the reasoning required by CoT $c$.

\newpage
\begin{table}[h!]
\centering
\scriptsize
\setlength{\tabcolsep}{2pt}
\renewcommand{\arraystretch}{0.76}
\begin{tabular}{@{}p{0.27\columnwidth}p{0.15\columnwidth}p{0.29\columnwidth}p{0.21\columnwidth}@{}}
\toprule
\textbf{Parameter} & \textbf{Value} &
\textbf{Parameter} & \textbf{Value} \\
\midrule
\multicolumn{2}{l}{\textit{Optimization and adaptation}} &
\multicolumn{2}{l}{\textit{On-policy rollout}} \\
Learning Rate & $5 \times 10^{-6}$ &
Max. Completion Length & 2,048 \\
Effective Batch Size & 32 &
Generations / Prompt & 1 \\
Optimizer & AdamW &
Sampling Temp. & 1.1 \\
Numerical Precision & bfloat16 &
Top-$p$ & 0.95 \\
Max. Gradient Norm & 0.1 &
Top-$k$ & 20 \\
LoRA Rank ($r$) & 64 &
Training Steps & \\
LoRA Alpha ($\alpha$) & 128 & & \\
\midrule
\multicolumn{4}{l}{\textit{RPT gate and anchoring}} \\
EMA Decay & 0.99 &
Gate Normalization & EMA z-score \\
Score Clipping & $[-5,5]$ &
Gate Sharpness ($\beta$) & 2.0 \\
Gate Threshold ($\tau$) & 0.0 &
Min. Gate ($g_{\min}$) & 0.05 \\
Anchoring Coefficient ($\lambda$) & 0.2 &
Gate Warmup & 5\% uniform + 5\% interpolation \\
\bottomrule
\end{tabular}
\caption{Training hyperparameters used for RP-OPSD.}
\label{tab:rp_opsd_hyperparameters}
\end{table}

\section{Prompt Templates}
\label{app:prompt_templates}

Our student prompt, full-teacher prompt, and language-specific reasoning prefix follow COPSD~\citep{liu2026copsd}.
The ablated-teacher prompt retains the target-language problem, its English translation, and the same language-specific instruction and reasoning prefix as the full teacher, but removes the English reference solution and its associated solution-conditioned instruction.
Figure~\ref{fig:ablated_teacher_prompt} shows the resulting template, with Swahili (SWA) as an example.

\FloatBarrier

\section{Training Hyperparameters}
\label{app:training_hyperparameters}

Across model scales and target languages, we follow the COPSD optimization and
on-policy rollout configuration~\citep{liu2026copsd}, adding only the RPT gate
and anchoring parameters. Table~\ref{tab:rp_opsd_hyperparameters} lists all
hyperparameters.

\end{document}